\documentclass[10pt,twocolumn,letterpaper]{article}

\usepackage{iccv}              %

\definecolor{gray}{gray}{0.95}
\definecolor{green}{HTML}{009000} 
\definecolor{red}{HTML}{ea4335} 
\usepackage{amsmath}
\usepackage{multirow}
\usepackage{algorithm}
\usepackage{algorithmic}

\newcommand{\hlg}[1]{\textcolor{green}{#1}}
\newcommand{\hlr}[1]{\textcolor{red}{#1}}
\newcommand{\better}[1]{\hlg{$\uparrow\,$#1}}
\newcommand{\worse}[1]{\hlr{$\uparrow\,$#1}}
\newcommand{\betterinv}[1]{\hlg{$\downarrow\,$#1}}
\newcommand{\worseinv}[1]{\hlr{$\downarrow\,$#1}}

\DeclareMathOperator{\gumbelsigmoid}{Gumbel-Softmax}
\DeclareMathOperator{\crossentropy}{Cross-Entropy}
\DeclareMathOperator{\kldivergence}{KL-Divergence}
\DeclareMathOperator{\forwardblock}{Forward-Block}
\DeclareMathOperator{\backwardblock}{Backward-Block}

\newlength\savewidth\newcommand\shline{\noalign{\global\savewidth\arrayrulewidth
  \global\arrayrulewidth 1pt}\hline\noalign{\global\arrayrulewidth\savewidth}}
\newcommand\hshline{\noalign{\global\savewidth\arrayrulewidth
  \global\arrayrulewidth 0.5pt}\hline\noalign{\global\arrayrulewidth\savewidth}}

\renewcommand{\paragraph}[1]{\vspace{1mm}\noindent\textbf{#1}}

\newcommand{\tablestyle}[2]{\setlength{\tabcolsep}{#1}\renewcommand{\arraystretch}{#2}\centering\footnotesize}

\definecolor{iccvblue}{rgb}{0.21,0.49,0.74}
\usepackage{hyperref}
\hypersetup{pagebackref,breaklinks,colorlinks,allcolors=iccvblue}

\def\paperID{11705} %
\def\confName{ICCV}
\def\confYear{2025}

\title{Dynamic Vision Mamba}

\author{
    Mengxuan Wu$^{1}$\footnotemark[1]\ , Zekai Li$^{1}$\footnotemark[1]\ \footnotemark[2]\ , Zhiyuan Liang$^{1}$\footnotemark[1]\ , Moyang Li$^{2}$, Xuanlei Zhao$^{1}$, Samir Khaki$^{3}$, Zheng Zhu$^{4}$, \\ Xiaojiang Peng$^{5}$, Konstantinos N. Plataniotis$^{3}$, Kai Wang$^{1}$\footnotemark[3]\ , Wangbo Zhao$^{1}$\footnotemark[3]\ , Yang You$^{1}$
    \\[0.04cm]
    $^{1}$NUS,
    $^{2}$ETH,
    $^{3}$University of Toronto,
    $^{4}$Tsinghua University,
    $^{5}$Shenzhen Technology University
}

\begin{document}
\maketitle
\renewcommand{\thefootnote}{\fnsymbol{footnote}}
\footnotetext[1]{equal contribution.}
\footnotetext[2]{project lead.}
\footnotetext[3]{corresponding author.}
\begin{abstract}
Mamba-based vision models have gained extensive attention as a result of being computationally more efficient than attention-based models. 
However, spatial redundancy still exists in these models, represented by token and block redundancy.
For token redundancy, we analytically find that early token pruning methods will result in inconsistency between training and inference or introduce extra computation for inference.
Therefore, we customize token pruning to fit the Mamba structure by rearranging the pruned sequence before feeding it into the next Mamba block.
For block redundancy, we allow each image to select SSM blocks dynamically based on an empirical observation that the inference speed of Mamba-based vision models is largely affected by the number of SSM blocks.
Our proposed method, Dynamic Vision Mamba (DyVM), effectively reduces FLOPs with minor performance drops.
We achieve a reduction of 35.2\% FLOPs with only a loss of accuracy of 1.7\% on Vim-S.
It also generalizes well across different Mamba vision model architectures and different vision tasks.
Our code will be made public at \href{https://github.com/NUS-HPC-AI-Lab/DyVM}{https://github.com/NUS-HPC-AI-Lab/DyVM}.
\end{abstract}

\section{Introduction}
\label{sec:intro}

\begin{figure*}[t]
    \centering
    \begin{subfigure}{0.68\textwidth}
        \centering
         \includegraphics[width=\textwidth]{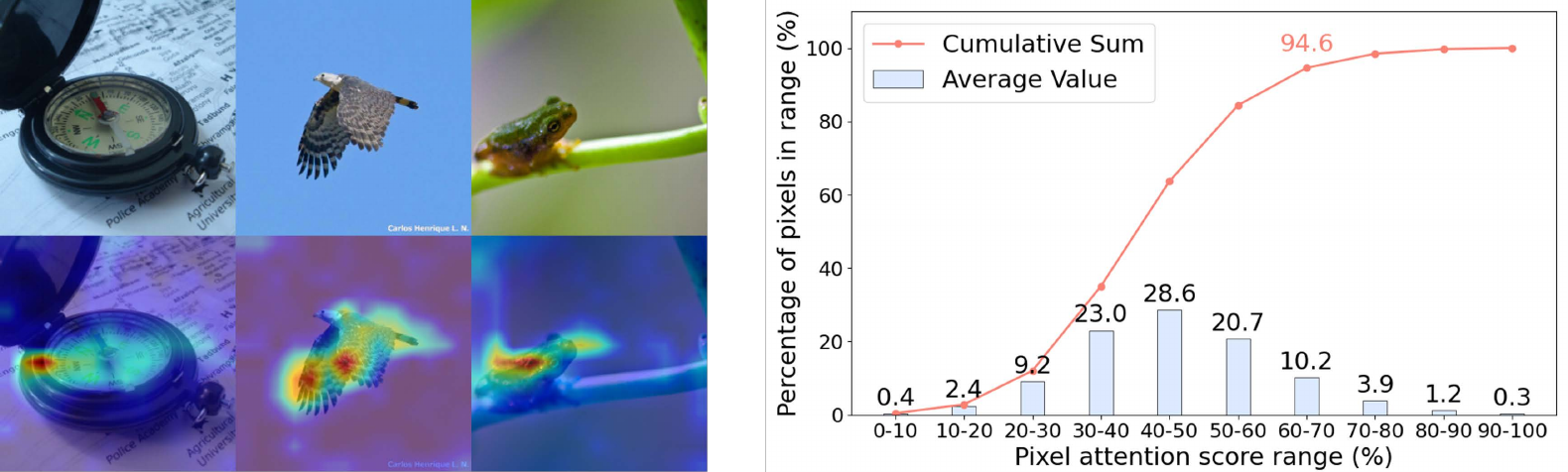}
         \caption{Mamba attention visualization and attention score statistics by HiddenMambaAttn~\cite{ali2024hiddenattentionmambamodels}, revealing significant token-level redundancy. 95\% of pixels have attention score less than 70\%, suggesting their minimal contribution to the model's performance.}
    \end{subfigure}
    \hfill
    \begin{subfigure}{0.30\textwidth}
        \centering
         \includegraphics[width=\textwidth]{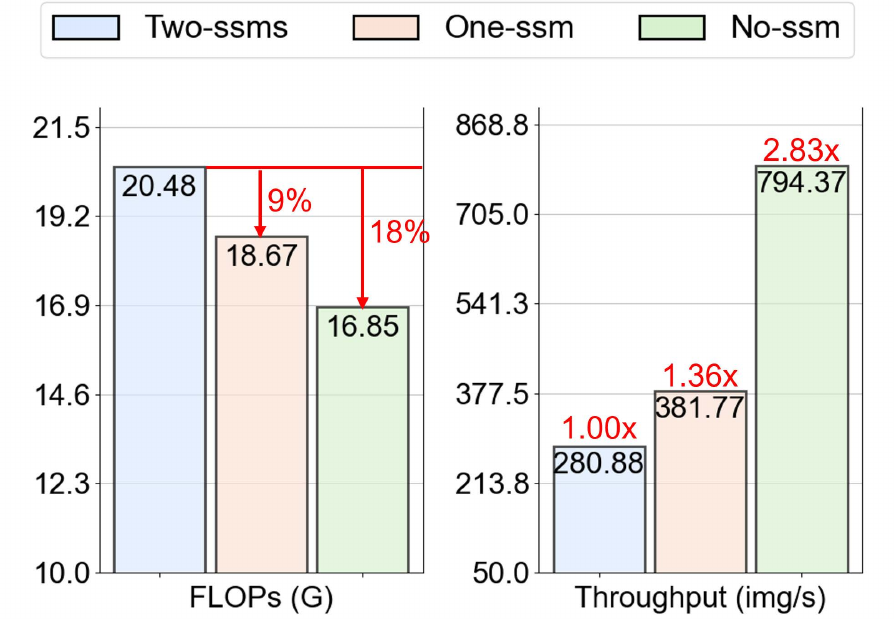}
         \caption{The inference speed is largely affected by the number of SSM blocks, as the significant throughput increment suggests.}
    \end{subfigure}
    \caption{\textbf{(a)} Pixel-wise attention score statistics computed from 1,000 images, with 10 images randomly sampled per class across 100 classes in ImageNet-1K dataset. \textbf{(b)} FLOPs and throughput performance under different SSM block number settings in Vim.}
    \label{fig:redundancy}
\end{figure*}

Vision Mambas~\cite{yang2024vivimvideovisionmamba, Liu2024SwinUMambaMU, Guo2024MambaIRAS, Pei2024EfficientVMambaAS} have gained promising performance on vision tasks, such as image classification~\cite{Vim, VMamba}, video understanding~\cite{VideoMamba}, and image segmentation~\cite{UMambaEL, Wang2024MambaUNetUP, Xing2024SegMambaLS, Ruan2024VMUNetVM}. Its key insight is to model the interaction between visual tokens with State Space Models (SSMs)~\cite{Mamba}.

Spatial redundancy, which has been proven widely to exist in Vision Transformers (ViTs)~\cite{Han2021DynamicNN, Wang2021NotAI, Meng2021AdaViTAV, Liang2022NotAP, Song2023DynamicGE}, may also be present in Vision Mambas.  
This redundancy appears on the token level, as \textit{representing an image with an excessive number of visual tokens}~\cite{Liu2023RevisitingTP, wei2023jointtokenpruningsqueezing}, thereby increasing computational cost and hindering inference speed. In Figure~\ref{fig:redundancy} (a), we randomly pick 10 images per class across 100 classes in the ImageNet-1K dataset and compute the attention score for each pixel. The statistics show 94.6\% of all pixels have attention scores less than 70\%, indicating minimal contribution to model performance. Although this issue has been sufficiently discussed and effectively resolved in ViT~\cite{ChannelPF, DynamicViT}, it remains inadequate for Vision Mambas.

To address the excessive number of visual tokens, token reduction has been proven to be an efficient solution in ViT scenarios. 
By masking out attention scores of undesired tokens, we can simulate token pruning during training and achieve consistency between training and inference.

However, simple masking method is not compatible with Mamba-based models.
To demonstrate this, we take Vim~\cite{Vim}, a representative Vision Mamba, as an example. 
When we attempt to mask out pruned tokens, illustrated in Figure~\ref{fig:different arch} (a), it leads to inconsistency of the output representation between training and inference, undermining the model's performance after visual token pruning. 
The failure can be attributed to Mamba's recurrent-like structure, where \textit{information from previous states is propagated through hidden states, and a simple masking disrupts this process}.

In addition to redundancy at the token level, we also notice the throughput bottleneck caused by multiple SSM blocks. For instance, Vim~\cite{Vim} implements both forward and backward SSMs at each layer to enhance spatial awareness. In Figure~\ref{fig:redundancy} (b), we compare the computational cost and inference throughput of Vim with two blocks, a single block, and no block at each layer. It can be observed that, although reducing SSM blocks has a marginal effect on FLOPs, it significantly increases the inference throughput, achieving a $1.36\times$ improvement with one block removed and a $2.83\times$ speedup with both blocks removed.  This finding corroborates our hypothesis that excessive SSM blocks in Vision Mambas impair efficiency. Consequently, identifying and deactivating those redundant blocks during inference is crucial.

Based on the above analysis, we introduce \textbf{Dy}namic \textbf{V}ision 
\textbf{M}amba (\textbf{DyVM}), designed to reduce redundancy at both token and block levels. From the token perspective, DyVM employs predictors at specific layers to identify and prune less informative tokens. To mitigate training-inference inconsistency, we rearrange pruned tokens to follow preserved tokens after masking during training, ensuring that preserved tokens remain unaffected by pruned tokens, as in the inference phase. Additionally, at the block level, DyVM dynamically selects the appropriate number of SSM blocks to process each image. This data-dependent approach specifically improves throughput for each sample.

Extensive experiments demonstrate that DyVM significantly reduces the FLOPs of Vision Mambas across various sizes while maintaining performance with only marginal decreases. We achieve a 35.2\% FLOPs reduction with only 1.7\% accuracy loss on Vim-S. Compared with visual token pruning baselines, \ie, HiddenAlign~\cite{HiddenAlign}, DyVM exhibits an improved performance-efficiency trade-off. Furthermore, experiments on VideoMamba~\cite{VideoMamba} and MambaReg~\cite{wang2024mambarvisionmambaneeds} verify the generalization ability of our method.

\section{Related Work}
\label{sec:related_work}

\paragraph{Mamba for Vision.}
The advancement of sequence modeling has significantly influenced computer vision, with models like ViT~\cite{ViT} adapted for vision tasks. Recently, State Space Models (SSMs) such as Mamba~\cite{Mamba} have gained attention for handling long sequences effectively. Follow-up works~\cite{SiMBA, MambaND, LocalMamba, MambaVision, MultiScale} have achieved strong performance on vision benchmarks using Mamba-based backbones. Vim~\cite{Vim} processes image patches with position embeddings and Mamba blocks integrating bidirectional SSMs. VMamba~\cite{VMamba} addresses Mamba's 1-D limitation by cross-scanning patches in four directions for 2-D dependencies. PlainMamba~\cite{PlainMamba} enhances feature fusion and generalization using zigzag scanning and direction-aware updates.

\paragraph{Token Pruning.}
Token pruning reduces computational cost by removing less important tokens, speeding up inference with minimal architectural changes. For ViTs~\cite{ViT}, methods like EViT~\cite{Liang2022NotAP} use class token attention scores, DynamicViT~\cite{DynamicViT} employs predictor layers, ToMe~\cite{ToMe} merges similar tokens, PATCHMERGER~\cite{renggli2022learnmerge} introduces a merger module, and T2T-ViT~\cite{T2TViT} aggregates neighboring tokens. However, these methods are unsuitable for vision Mamba due to structural differences. HiddenAlign~\cite{HiddenAlign} explores token-level pruning in Mamba but incurs extra inference costs. In contrast, our method combines token and block-level pruning, achieving comprehensive improvements without additional computation.

\section{Method}
\label{sec:method}

\begin{figure}[t]
    \centering
    \begin{subfigure}{0.45\textwidth}
        \centering
        \includegraphics[width=\textwidth]{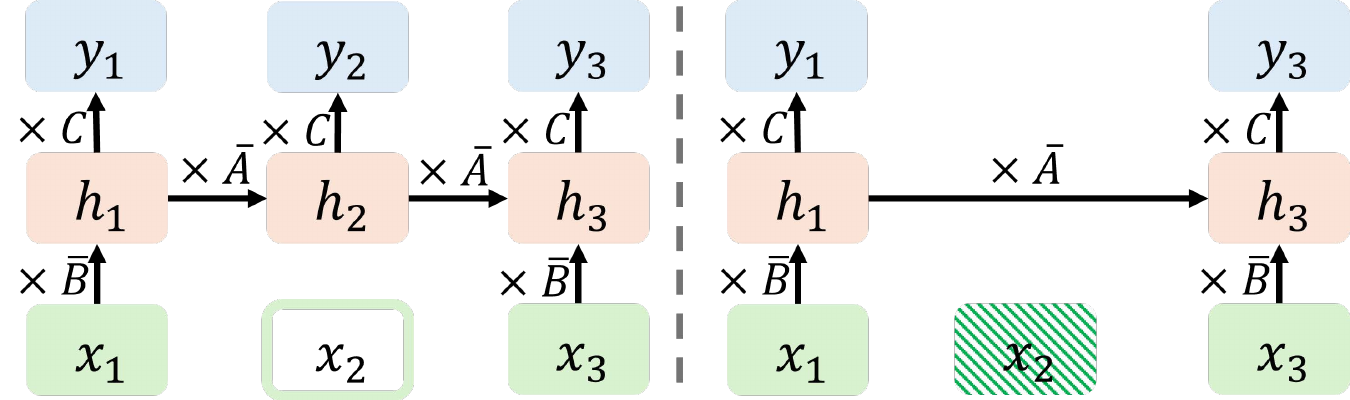}
        \caption{Plain token masking method causes inconsistency between training and inference, primarily as different number of evolution transformations (\ie, $\bar{A}$).}
    \end{subfigure}
    \begin{subfigure}{0.45\textwidth}
        \centering
        \includegraphics[width=\textwidth]{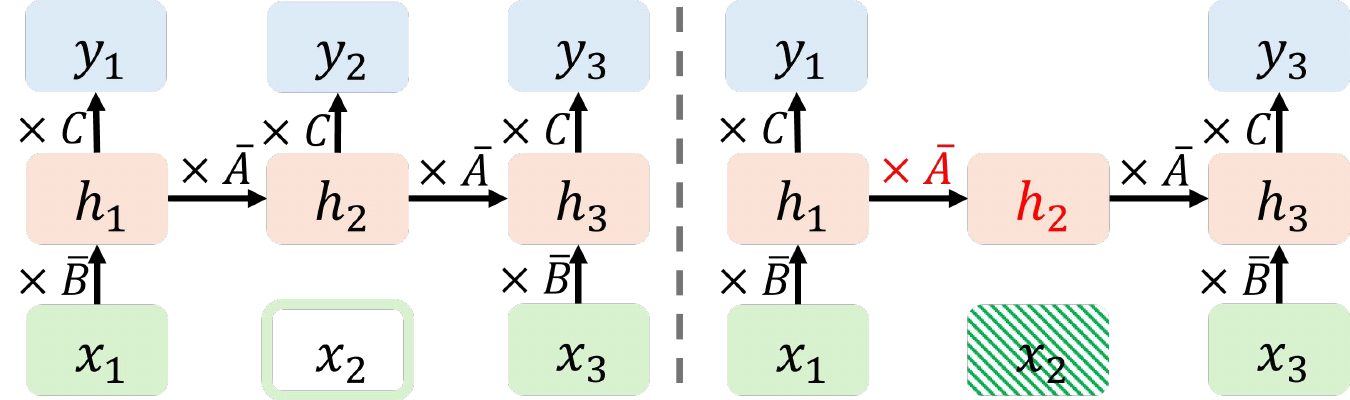}
        \caption{HiddenAlign~\cite{HiddenAlign} retains evolution transformations (\ie, $\bar{A}$) for pruned tokens during inference, thus achieving consistency between training and inference at the cost of extra computation.}
    \end{subfigure}
    \begin{subfigure}{0.45\textwidth}
        \centering
        \includegraphics[width=\textwidth]{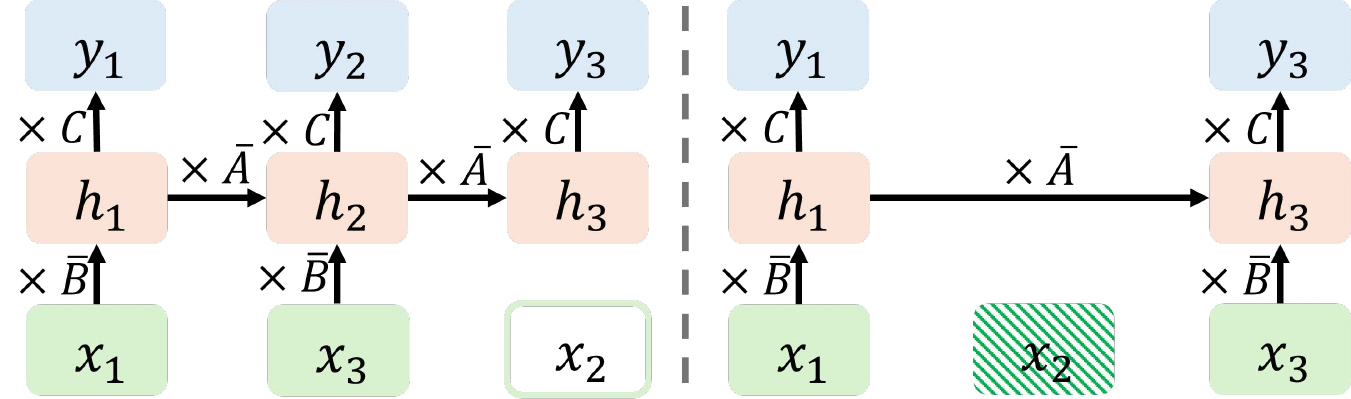}
        \caption{DyVM rearranges tokens during training to achieve consistency, eliminating extra computation during inference comparing with HiddenAlign.}
    \end{subfigure}
    \caption{Demonstration of three token pruning methods' training (left) and inference (right), with a 3-token sequence example where the middle token is pruned. Solid fill indicates retained tokens, unfilled elements represent masked tokens during training, and diagonal line patterns denote tokens dropped during inference.}
    \label{fig:different arch}
\end{figure}

\subsection{Preliminary}
\label{preliminary}
State Space Models (SSMs)~\cite{Gu2020HiPPORM, Gu2021EfficientlyML, Gu2022OnTP} map an input sequence $x(t) \in \mathbb{R}^{L}$ to an output sequence $y(t) \in \mathbb{R}^{L}$ by propagating information through hidden states $h(t) \in \mathbb{R}^{N}$:
\begin{align}
\label{eq1}
    h'(t) &= A h(t) + B x(t), \\ 
    y(t)  &= Ch'(t).
\end{align}
$A \in \mathbb{R}^{N\times N}$ are evolution parameters and $B \in \mathbb{R}^{L \times N}$ and $C \in \mathbb{R}^{N\times L}$ are projection parameters.

While SSM targets continuous input, Mamba~\cite{Mamba} provides a discrete version by introducing a timescale parameter $\Delta$ with zero-order hold (ZOH):
\begin{align}
    \bar{A} &= \exp (\Delta A), \\ 
    \bar{B} &= (\Delta A)^{-1}(\exp(\Delta A)-I)\Delta B.
\end{align}
Correspondingly, the discrete version of equation~\ref{eq1} can be formulated as:
\begin{align}
    h_t &= \bar{A} h_{t-1} + \bar{B} x_t, \\ 
    y_t &= Ch_t.
\end{align}
The Mamba model applies a global convolution to compute the output as follows:
\begin{align}
    \bar{K} &= (C\bar{B}, C\bar{A}\bar{B}, \dots, C\bar{A}^{L-1}\bar{B}), \\ 
    y &= x * \bar{K}.
\end{align}
To handle 2D images, Vision Mambas~\cite{Vim, LocalMamba, VideoMamba, VMamba} transform it into a sequence of tokens, as that is done in vision transformers~\cite{ViT}. Subsequently, a series of mamba layers with SSMs are adopted to build the relationship between tokens. Different approaches vary in the design of layers.

\subsection{Existing Masking Method}
Existing token pruning methods~\cite{DynamicViT} for vision transformers use masks at training time to simulate the removal of tokens, yet these approaches are not directly compatible with the Mamba-based model structure.

\begin{figure*}[t]
    \centering
    \includegraphics[width=0.9\textwidth]{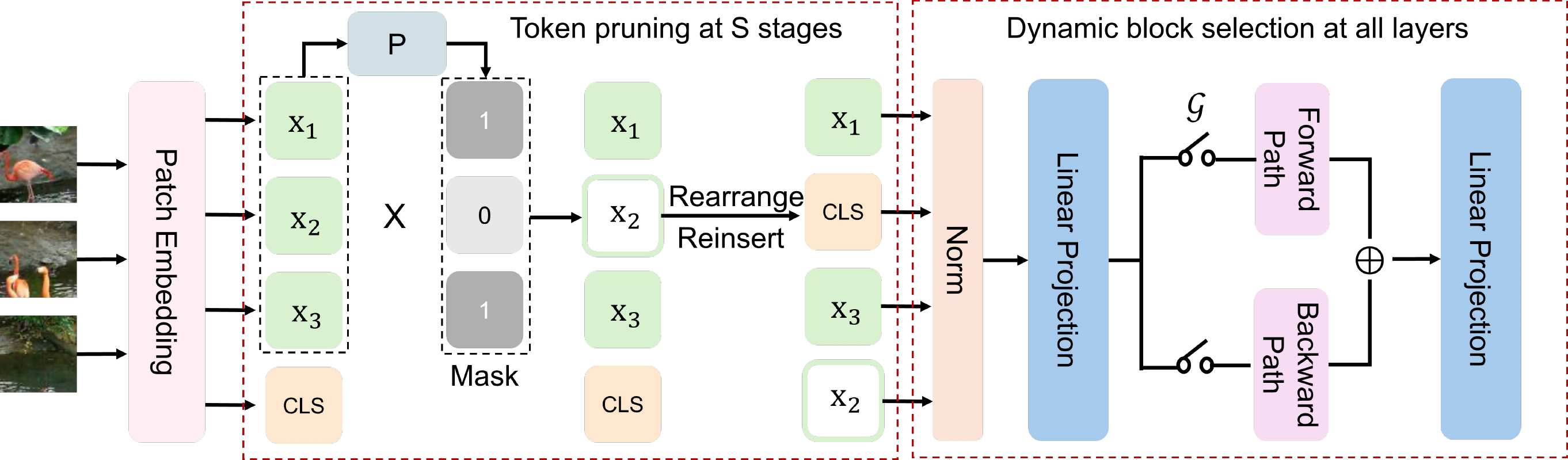}
    \caption{Dynamic Vision Mamba pipeline. The predictor modules are inserted between specific mamba blocks to gradually prune redundant tokens, while block selection modules are embedded into every mamba block to select SSM blocks for each sample dynamically. With these two methods, DyVM greatly reduces the FLOPs of the model.}
    \label{fig2}
\end{figure*}

\paragraph{Plain token masking}
The most direct method is to mask tokens by setting their embeddings to zero during training, as displayed in Figure~\ref{fig:different arch} (a).
Although it blocks certain tokens' information while allowing the remaining tokens' information to propagate, it results in inconsistency during training and inference, undermining the model's performance.
Below, we provide a detailed analysis.
For a sequence of $L$ input tokens $X \in \mathbb{R}^{L \times D}$, with $K$ tokens retained while the other tokens are pruned, let $\{d_i\}_0^{K-1}$ denote the indices of retained tokens ($d_i < d_j$ if $i < j$).
With masking tokens, the output of each sequence position of the Mamba block during training can be computed as follows:
\begin{equation}
        y_{d_{i}} = C(\textcolor{red}{\bar{A}^{d_{i} - d_0}}\bar{B}x_{d_0} + \textcolor{red}{\bar{A}^{d_{i} - d_1}}\bar{B}x_{d_1} + \cdots + \bar{B}x_{d_{i}}).
    \label{eq:in-place masking}
\end{equation}
During inference, redundant tokens are directly dropped, and retained tokens are concatenated.
Suppose we retain the tokens with the same indices, then the output of each sequence position of the Mamba block during inference can be computed as follows:
\begin{equation}
        y_{d_{i}}' = C(\textcolor{red}{\bar{A}^{i}}\bar{B}x_{d_0} + \textcolor{red}{\bar{A}^{i-1}}\bar{B}x_{d_1} + \cdots + \bar{B}x_{d_{i}}).
    \label{eq:inference}
\end{equation}
The training and inference outputs are highly inconsistent in the number of evolution transformations (\ie, $\bar{A}$), since $d_{i} - d_0 \geq i$. Only in rare instances where $\{d_i\}_{0}^{i}$ are consecutive indices in the training sequence, can the equality be achieved.

\paragraph{HiddenAlign~\cite{HiddenAlign}}
Previous work HiddenAlign (HA) is aware of such inconsistencies in early methods and proposes a new approach, as displayed in Figure~\ref{fig:different arch} (b). 
During training, HA uses the same token masking method as in the plain token masking method. 
During inference, for each of the pruned tokens, HA retains its evolution transformation (\ie, $\bar{A}$), while pruning its corresponding projections (\ie, $\bar{B}$, and $C$). 
With this new approach, both the training and inference outputs are consistent and computed as:
\begin{equation}
        y_{d_{i}} = y_{d_{i}}' = C(\textcolor{red}{\bar{A}^{d_{i} - d_0}}\bar{B}x_{d_0} + \textcolor{red}{\bar{A}^{d_{i} - d_1}}\bar{B}x_{d_1} + \cdots + \bar{B}x_{d_{i}}).
    \label{eq:HA}
\end{equation}
However, compared with the plain token pruning method, this approach introduces extra computation during inference. In most cases, HA's inference cost (Equation~\ref{eq:HA}) is greater than the plain token masking method's inference cost (Equation~\ref{eq:inference}), since $d_{i} - d_{0} \geq i$, and equality is rarely achieved if and only if $\{d_i\}_{0}^{K-1}$ are consecutive indices.

\subsection{Dynamic Vision Mamba}

Our analysis raises a key question: Can token pruning achieve training-inference consistency without extra computational overhead? 
We propose Dynamic Vision Mamba (DyVM), which reduces Vision Mamba's spatial redundancy at both token and block levels. 
The token pruning is performed gradually with $S$ pruning stages, with each continuing to mask out tokens based on the previous stage.
In each stage, tokens are rearranged during training to mimic the ordering during inference, thus achieving consistency while eliminating extra computations.
The dynamic block \footnote{In DyVM, a block consists of a 1-D causal convolution and a selective scanning module.} selection is performed at every layer, predicting which block(s) each sample should pass through.
We illustrate DyVM's pipeline in Figure~\ref{fig2}.

\paragraph{Token Pruning} \label{token_pruning}
At each stage $s$, we prune a fixed rate of tokens and proceed by maintaining a binary mask $M^{s} \in \{0, 1\}^{B \times L}$ for each token, indicating whether to retain it or drop it.
All elements in $M^0$ are initialized to 1.
Following DynamicViT~\cite{DynamicViT}, we implement a predictor $P$ followed by softmax at each pruning stage to generate the probability of pruning and retaining each token for the batched input sequences $H \in \mathbb{R}^{B \times L \times D}$:
\begin{equation}
    \Pi = \text{Softmax}(P(H, M^{s-1})) \in \mathbb{R}^{B \times L \times 2}.
\end{equation}
where $\pi_{b, i, 0}$ is the probability of retaining the $i$-th token in the $b$-th batch, and $\pi_{b, i, 1}$ is the probability of pruning it.
Then, $M^s$ is updated by the current policy $\hat{M}$:
\begin{align}
    \hat{M} &= \gumbelsigmoid(\Pi) \in \{0, 1\}^{B \times L}, \\
    M^s &= \hat{M} \odot M^{s-1}.
\end{align}
Here, we adopt the gumbel-softmax~\cite{jang2016categorical} trick to make the sampling process differentiable, therefore enabling the end-to-end training. The pruning effect is achieved by multiplying tokens by the updated mask $M$\footnote{When masking the sequence, the class token will always be preserved.}.

To solve the training-inference inconsistency issue mentioned above, we propose to rearrange the positions of tokens before passing them to the SSM block during training, as displayed in Figure~\ref{fig:different arch} (c).
Specifically, for each sequence $X \in \mathbb{R}^{L \times D}$, we begin by splitting class tokens $c \in \mathbb{R}^{1 \times D}$ and the other tokens.
We only operate on the sequences without class tokens.
Let $\{d_i\}_0^{K-1}$ denote the indices of $K$ retained tokens ($d_i < d_j$ if $i < j$), given by $M^s$.
Instead of masking tokens in place, we aggregate retained tokens into one contiguous block while preserving their relative order.
Then, we reinsert the class token $c$ back to the middle, which is the commonly used class token position in the Vim model:
\begin{equation}
    X^{\text{retained}} = [x_{d_0}, \dots, c, \dots x_{d_{K-1}}] \in \mathbb{R}^{(K+1) \times D},
\end{equation}
Similarly, we group pruned tokens into another contiguous block.
Let $\{p_i\}_0^{N-K-2}$ denote indices of pruned tokens ($p_i < p_j$ if $i < j$).
The pruned token block is:
\begin{equation}
    X^{\text{pruned}} = [x_{p_0}, x_{p_1}, \dots, x_{p_{N-K-2}}] \in \mathbb{R}^{(N-K-1) \times D},
\end{equation}
Finally, we concatenate retained and pruned blocks:
\begin{equation}
    X^{\text{rearranged}} = \text{Concat}(X^{\text{retained}}, X^{\text{pruned}}) \in \mathbb{R}^{L \times D}.
\end{equation}
This formulation removes the problem of unintended information propagation through hidden states, maintaining the consistency between training and inference and improving the model’s performance and stability. Now both the training and inference outputs are consistent and computed as:
\begin{equation}
        y_{i} = y_{d_{i}}' = C(\textcolor{red}{\bar{A}^{i}}\bar{B}x_{d_0} + \textcolor{red}{\bar{A}^{i - 1}}\bar{B}x_{d_1} + \cdots + \bar{B}x_{d_{i}}).
\end{equation}

\paragraph{Dynamic block Selection}\label{block_selection}
In Figure~\ref{fig:redundancy}(b), we show that the throughput decreases as the number of active scanning blocks increases.
Thus, we propose to dynamically select scanning blocks for each sample so that redundancy is further reduced at the block level.
Specifically, a sample could pass through both forward and backward blocks, one of the forward and backward blocks, or even none of the two blocks.
This is achieved by a block selector within each Vim layer, which predicts scores of each block given class tokens $C^l \in \mathbb{R}^{B \times D}$ at layer $l$ as input.
This is followed by a Gumbel-sigmoid function that transforms the score matrix into a binary mask:
\begin{equation}
    Q^{l} = \gumbelsigmoid(\mathcal{G}(C^l)) \in \mathbb{R}^{B \times 2},
\end{equation}
Finally, the redundant block for each sample is deactivated by multiplying the outputs with the masks as follows:
\begin{align}
    O^{l, f} &= \forwardblock(H^l) \cdot Q^{l}_{:,0} \in \mathbb{R}^{B \times L \times D}, \\
    O^{l, b} &= \backwardblock(H^l) \cdot Q^{l}_{:,1} \in \mathbb{R}^{B \times L \times D}.
\end{align}
$H^l \in \mathbb{R}^{B \times L \times D}$ are \textbf{rearranged} input sequences of the forward and backward blocks at layer $l$. 
$O^{l, f}, O^{l, b}$ denote masked outputs of forward and backward blocks at layer $l$, respectively.
Notice that it is not feasible to directly apply masks on forward or backward input because of non-zero bias terms in each block.

\begin{table*}[t]
    \centering
    \tablestyle{10pt}{1.3}
    \begin{tabular}{c|cccccc}
        model & params (M) & FLOPs (G) & $\Delta_{\text{FLOPs}}$ & top-1 acc. (\%) & $\Delta_{\text{acc}}$\\
        \shline
        Vim-T               & 7     & 1.45 & / & 76.1 & / \\
        Vim-S               & 26    & 5.08 & / & 80.5 & / \\
        Vim-B               & 98    & 18.87 & / & 81.9 & / \\
        Vim-T + HA          & 7     & 1.29 & \betterinv{11.0\%}  & 75.1 & \worseinv{1.0} \\
        Vim-S + HA          & 26    & 3.60 & \betterinv{29.1\%}  & 78.8 & \worseinv{1.7} \\ \rowcolor{gray}
        Vim-T + DyVM        & 7     & 1.25 & \betterinv{13.8\%}  & 75.2 & \worseinv{0.9} \\ \rowcolor{gray}
        Vim-S + DyVM        & 27    & 3.29 & \betterinv{35.2\%}  & 78.8 & \worseinv{1.7} \\ \rowcolor{gray}
        Vim-B + DyVM        & 101   & 12.09 & \betterinv{35.9\%} & 80.0 & \worseinv{1.9} \\ 
        \hshline
        VideoMamba-T        & 7     & 1.45 & / & 76.9 & / \\ 
        VideoMamba-S        & 26    & 5.08 & / & 81.2 & / \\  \rowcolor{gray}
        VideoMamba-T + DyVM & 7     & 1.22 & \betterinv{15.9\%} & 75.0 & \worseinv{1.9} \\ \rowcolor{gray}
        VideoMamba-S + DyVM & 27    & 3.75 & \betterinv{26.2\%} & 79.9 & \worseinv{1.3} \\ 
        \hshline
        MambaReg-S          & 29 & 5.36 & / & 81.4 & / \\
        MambaReg-B          & 98 & 19.92 & / & 83.0 & / \\ \rowcolor{gray}
        MambaReg-S + DyVM   & 29 & 3.57 & \betterinv{33.4\%} & 79.3 & \worseinv{2.1} \\ \rowcolor{gray}
        MambaReg-B + DyVM   & 102 & 13.14 & \betterinv{34.0\%} & 80.5 & \worseinv{2.5} \\
    \end{tabular}
    \caption{Results of image classification on ImageNet-1K. Compared with HiddenAlign (HA), DyVM demonstrates higher FLOP reduction with similar test accuracy on Vim-T and Vim-S. On Vim-B, we achieve much lower FLOPs with a minor performance drop. DyVM also generalizes well on VideoMamba and MambaReg under the classification setting. }
    \label{tab:ImageClassification}
\end{table*}

\subsection{Training and Inference}
\paragraph{Training.} The training objectives of DyVM consist of five components: one classification loss, two supervision losses to constrain the pruning ratio, and two distillation losses to calibrate model performance.

Firstly, we compute the standard cross-entropy loss between model predictions $\hat{y}$ and ground truth labels $y$ as classification loss:
\begin{equation}
    \mathcal{L}_{cls} = \crossentropy(\hat{y}, y).
\end{equation}

Secondly, to supervise the token pruning ratio, we set a target token ratio $\rho$, and expect to retain $\lfloor \rho^i L \rfloor$ tokens after the $i$-th pruning stage.
Given a set of $S$ pruning stages with target ratios $\boldsymbol{\rho} = [\rho, \rho^2, \rho^3, \dots, \rho^S]$, we calculate a MSE loss:
\begin{equation}
    \mathcal{L}_{token} = \frac{1}{BS}\sum_{b=1}^B \sum_{s=1}^S \left(\rho^s - \frac{1}{L} \sum_{i=1}^L M_{b, i}^{s}\right)^2.
    \label{eq:token}
\end{equation}
where $\hat{M}^{b, s}_i$ denotes the $i$-th value of the mask of batch $b$ after the pruning stage $s$.

To supervise the block selection ratio, we compute the average ratio of active blocks across all layers ($N$ in total) and calculate an MSE loss with a predefined block ratio $\rho^{p}$:
\begin{equation}
    \mathcal{L}_{block} = \left(\rho^{p} - \frac{1}{BN}\sum_{i=1}^{B}\sum_{j=1}^{N} \frac{Q^{j}_{i, 0}+Q^{j}_{i, 1}}{2}\right)^2.
    \label{eq:block}
\end{equation}

Lastly, we further calibrate the model's behavior after token pruning and block selection by using the original backbone network as a teacher model.
Firstly, we minimize the Kullback-Leibler (KL) divergence loss between the model's outputs $\hat{y}$ and a teacher model's output $y^{*}$:
\begin{equation}
    \mathcal{L}_{dis\_out} = \kldivergence(\hat{y}||y^{*}).
    \label{eq:DisOut}
\end{equation}
Additionally, we make all retained tokens close to those from the teacher model by calculating an MSE loss:
\begin{equation}
    \mathcal{L}_{dis\_token} = \frac{1}{\sum_{b=1}^B \sum_{l=1}^L M^b_l} \sum_{b=1}^B \sum_{l=1}^L M^b_l (\hat{t}^b_l - {t^*}^b_l)^2.
    \label{eq:DisToken}
\end{equation}

The joint loss is a weighted sum of the above five losses:
\begin{align}
    \mathcal{L}_{joint} = &\lambda_{cls}\mathcal{L}_{cls} + \lambda_{token}\mathcal{L}_{token} + \lambda_{block}\mathcal{L}_{block} \notag \\ & + \lambda_{dis\_out}\mathcal{L}_{dis\_out} +\lambda_{dis\_token}\mathcal{L}_{dis\_token}.
\end{align}

\paragraph{Inference.}
During inference, pruned tokens are dropped and blocks skipped directly for higher efficiency.
For token pruning, given the target ratio $\rho$, we retain $K = \lfloor \rho^s N \rfloor$ tokens after the $s$-th pruning stage and drop the others.
The indices of retained tokens are obtained by sorting tokens by the retaining probability and choosing the top-$K$ tokens.
\begin{align}
    \mathcal{I} &= \text{argsort}(\pi_{:, 0}), \\
    \mathcal{I}^{t}_{retained} &= \mathcal{I}_{0:K-1}.
\end{align}

For block selection, taking the forward block as an example, only samples with block mask values $1$ are sent into the forward block at layer $l$. 
Formally, indices of samples passing through the forward block at layer $l$ are:
\begin{equation}
    \mathcal{I}^{p}_{\text{retained}} = \{1 \leq i \leq B: Q^{l}_{i, 0} = 1 \}.
\end{equation}
The backward block follows the same logic.
Consequently, fewer convolution and SSM scanning computations are made during evaluation, accelerating the inference process.

\begin{table}[t]
    \centering
    \tablestyle{4pt}{1.3}
    \begin{tabular}{c|cccc}
         model & FLOPs (G) & $\Delta_{\text{FLOPs}}$ & top-1 acc. (\%) & $\Delta_{\text{acc}}$ \\
         \shline
         VideoMamba-T & 11.34 & / & 76.9 & / \\
         VideoMamba-S & 40.03 & / & 79.3 & / \\ \rowcolor{gray}
         VideoMamba-T + DyVM & 7.02 & \betterinv{38.1\%} & 74.3 & \worseinv{2.6} \\ \rowcolor{gray}
         VideoMamba-S + DyVM & 25.35 & \betterinv{36.7\%} & 76.3 & \worseinv{3.0} \\
    \end{tabular}
    \caption{Results of video understanding on Kinetics-400. This demonstrates our method can be generalized to other modality, with similar impact on FLOPs and accuracy. We use 8 frames for input, and set target token ratio to 0.7 and block ratio to 0.8.}
    \label{tab:videoUnderstanding}
\end{table}

\begin{table}[t]
    \centering
    \tablestyle{4pt}{1.3}
    \begin{tabular}{c|cc}
         model & mIoU (\%) & $\Delta_{\text{mIoU}}$ \\
         \shline
         Vim-T & 41.0 & / \\
         Vim-S & 44.9 & / \\ \rowcolor{gray}
         Vim-T + DyVM & 40.1 & \worseinv{0.9} \\ \rowcolor{gray}
         Vim-S + DyVM & 42.0 & \worseinv{2.9} \\
    \end{tabular}
    \caption{Results of semantic segmentation on ADE20K. This demonstrates DyVM can adapt to dense prediction tasks. We set target token ratio to 0.9 for tiny model and 0.8 for small model, combined with a consistent block ratio of 0.8 across models.}
    \label{tab:segmentation}
\end{table}

\section{Experiment}

\begin{figure*}[ht]
    \centering
    \begin{subfigure}{0.42\textwidth}
        \centering
        \includegraphics[width=\textwidth]{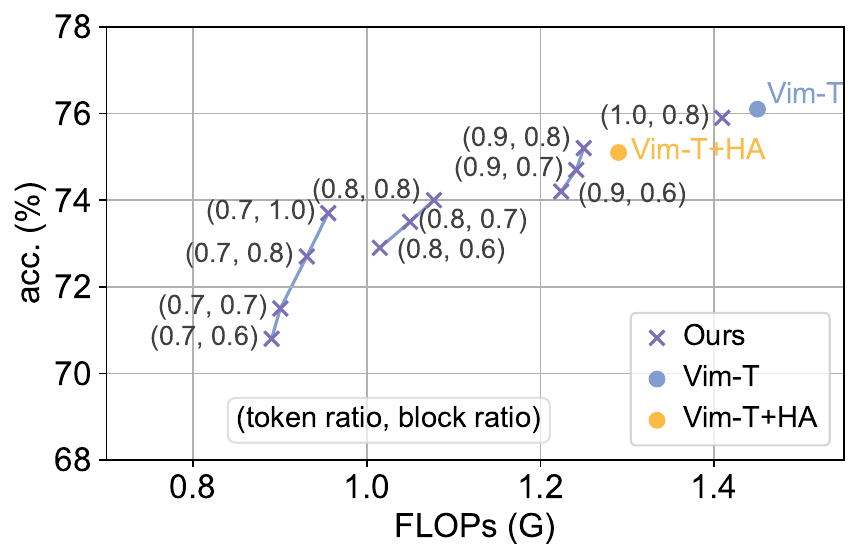}
        \caption{Results of different token-block ratio combinations on Vim-T.}
    \end{subfigure}
    \hspace{1em}
    \begin{subfigure}{0.42\textwidth}
        \centering
        \includegraphics[width=\textwidth]{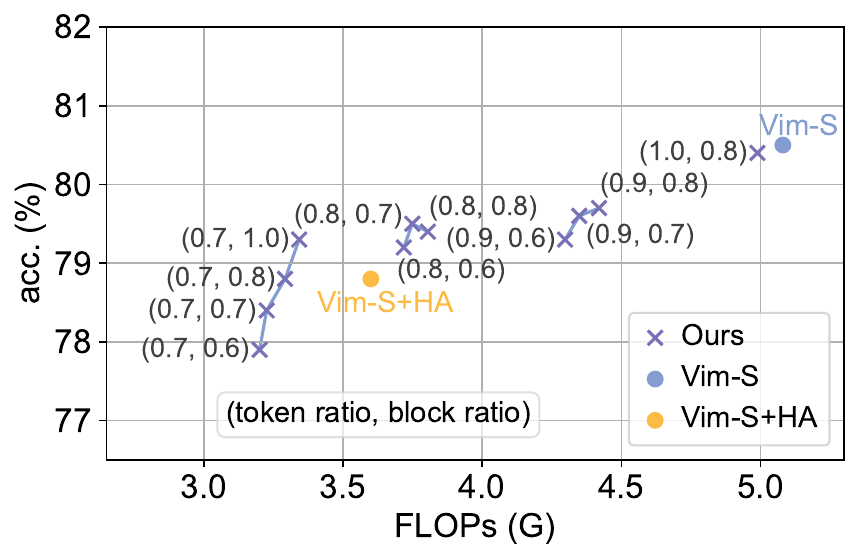}
        \caption{Results of different token-block ratio combinations on Vim-S.}
    \end{subfigure}
    \caption{The trade-off between FLOPs and accuracy represents a balance between model efficiency and performance. Larger models are less impacted by pruning, indicating they have greater spatial redundancy and can tolerate more aggressive pruning.}
    \label{fig:ratio-acc}
\end{figure*}

\begin{table*}[t]
    \centering
    \begin{subtable}[t]{0.4\textwidth}
        \centering
        \tablestyle{6pt}{1.2}
        \begin{tabular}{ccccc}
            model & \multicolumn{2}{c}{Vim-T} & \multicolumn{2}{c}{Vim-S} \\
            \shline
            strategy & top-1 acc. (\%) & $\Delta$ & top-1 acc. (\%) & $\Delta$ \\
            random & 70.2 & \worseinv{3.5} & 77.8 & \worseinv{1.5} \\
            static & 68.0 & \worseinv{5.7} & 77.4 & \worseinv{1.9} \\ \rowcolor{gray}
            learnable & 73.7 & /  & \textbf{79.3} & / \\ 
        \end{tabular}
        \caption{Learnable token mask brings the best performance.}
        \label{tab:DifferentTokenPruningStrategy}
    \end{subtable}
    \hfill
    \begin{subtable}[t]{0.5\textwidth}
        \centering
        \tablestyle{6pt}{1.2}
        \begin{tabular}{cc >{\columncolor{gray}}c cc >{\columncolor{gray}}c c}
            model & \multicolumn{3}{c}{Vim-T} & \multicolumn{3}{c}{Vim-S} \\
            \shline
            block ratio & rand. & ours &$\Delta$ & rand. & ours &$\Delta$           \\ 
            0.8 &69.4&74.9& \better{5.5} &77.2&80.3& \better{3.1} \\
            0.9 &72.8&76.0& \better{3.2} &78.5&\textbf{80.5}& \better{2.0} \\
        \end{tabular}
        \caption{Learnable predictors for block selection bring the best performance.}
        \label{tab:DifferentblockPruningStrategy}
    \end{subtable}
    \caption{Comparisons of different pruning strategies: (a) Different token pruning method (b) Different block selection method. For (a), we use token pruning with 0.7 target ratio and no block selection. For (b), we only use block selection.}
    \label{tab:Ablation1}
\end{table*}

\begin{table}[t]
    \tablestyle{6pt}{1.2}
    \begin{tabular}{ccccc}
        model & \multicolumn{2}{c}{Vim-T} & \multicolumn{2}{c}{Vim-S} \\
        \shline
        \# stage & top-1 acc. (\%) & $\Delta$ & top-1 acc. (\%) & $\Delta$ \\
        1 & 73.0 & \worseinv{0.7} & 74.6 & \worseinv{4.7}  \\
        2 & 73.6 & \worseinv{0.1} & 74.9 & \worseinv{4.4} \\ \rowcolor{gray}
        3 & 73.7 & / & \textbf{79.3} & / \\
    \end{tabular}
    \caption{Comparisons of different number of pruning stage. Gradually pruning tokens in three stages brings the best performance. We use token pruning with 0.7 target ratio and no block selection.}
    \label{tab:NumberOfPruningStage}
\end{table}

\subsection{Models and Dataset}

\paragraph{Models.}
We implement DyVM on Vim models (Vim-T, Vim-S, Vim-B)~\cite{Vim} and compare it with HiddenAlign~\cite{HiddenAlign} (HA) as a baseline. To demonstrate DyVM's generalization capability, we integrate it into VideoMamba (VideoMamba-T, VideoMamba-S)~\cite{VideoMamba} and MambaReg (MambaReg-S, MambaReg-B)~\cite{wang2024mambarvisionmambaneeds} for image classification. Furthermore, we conduct an assessment of DyVM's cross-modal generalization capability through the evaluation of the DyVM-integrated VideoMamba framework on video understanding tasks. Additionally, following HA's experimental setup, we evaluate DyVM on semantic segmentation using UperNet~\cite{xiao2018unifiedperceptualparsingscene} as the base framework.

\paragraph{Dataset.}
For the image classification task, we conduct experiments on ImageNet-1K~\cite{ImageNet}, which consists of 1281167 images categorized into 1000 classes.
For the video understanding task, we conduct experiments on Kinetics-400~\cite{kinetics400}, which covers 400 human action classes with 650000 videos.
For semantic segmentation, we conduct experiments on ADE20K~\cite{ade20k}, a large-scale dataset with 20000 images spanning 150 semantic categories.

\subsection{Experiment Settings}
For the image classification task, we train our model by fine-tuning the backbone model for 30 epochs.
We set the learning rate for the Tiny size model as 3e-5, and for the Small and Base sizes as 5e-5.
We use a cosine learning rate scheduler with a 5-epoch warm-up phase.
The batch sizes for the Tiny, Small, and Base size models are 128, 64, and 32, respectively.
For token pruning, we set $S = 3$ token pruning stages with a pruning rate $\boldsymbol{\rho} = [\rho, \rho^2, \rho^3]$, where $\rho$ is the target token ratio.
For block selection, we set one single target block ratio $r$ across all layers.
Specifically, we initialize the block selection module to preserve all samples, ensuring it closely mimics the original model's behavior.
We use $\lambda_{cls} = 1$, $\lambda_{token} = \lambda_{block} = 10$, and $\lambda_{dis\_out} = \lambda_{dis\_token} = 0.5$ when calculating the joint loss.
Other training settings and details can be found in the supplementary material.

\subsection{Main Results}

\paragraph{Comparison with baselines.}
In Table~\ref{tab:ImageClassification}, we present the results of applying DyVM to Vim under the image classification setting.
DyVM successfully reduces Vim's FLOPs on all model sizes while maintaining satisfactory performance.
Compared with the HA method, DyVM achieves the same or better performance on Vim-T and Vim-S with \textbf{larger} FLOPs reduction.
DyVM also generalizes well on VideoMamba and MambaReg, reducing considerable FLOPs with minor performance drops. 

\paragraph{Results of different ratio combinations.}
Figure \ref{fig:ratio-acc} shows top-1 accuracy and FLOPs for various token-block ratio combinations on Vim-T and Vim-S. Token pruning reduces FLOPs by shortening sequence length but causes greater accuracy loss as the ratio increases. Combining it with block selection achieves similar FLOPs reduction with less performance drop. Larger models, having more spatial redundancy, tolerate aggressive pruning better, showing less performance degradation.

\paragraph{Scaling to larger token number}
In Table~\ref{tab:videoUnderstanding}, we present the results of DyVM applied to VideoMamba on the K-400 video-understanding dataset.
We consistently reduce considerable FLOPs while maintaining comparable performance.
This shows that DyVM is robust under a larger token number setting.

\paragraph{Results on semantic segmentation.}
DyVM demonstrates strong adaptability to prediction tasks when integrated into UperNet for semantic segmentation on ADE20K. As shown in Table~\ref{tab:segmentation}, the framework maintains competitive segmentation accuracy (mIoU) while reducing computational costs.

\begin{figure*}[t]
    \centering
    \includegraphics[width=0.9\textwidth]{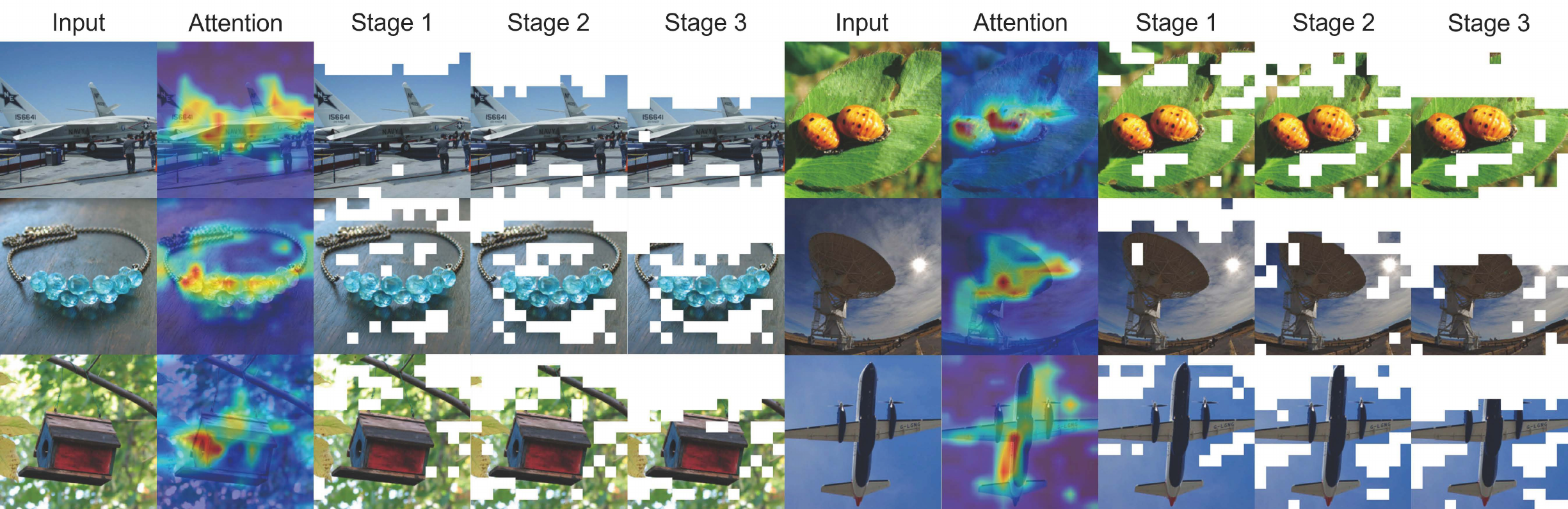}
    \caption{Visualization of token pruning results. In each group of images, we show the original image, along with its hidden attention and retained tokens of each pruning stage. Pruned tokens are mostly from low-attention areas, implying their redundancy.}
    \label{TokenVisual}
\end{figure*}

\begin{figure*}[t]
    \centering
    \includegraphics[width=0.9\textwidth]{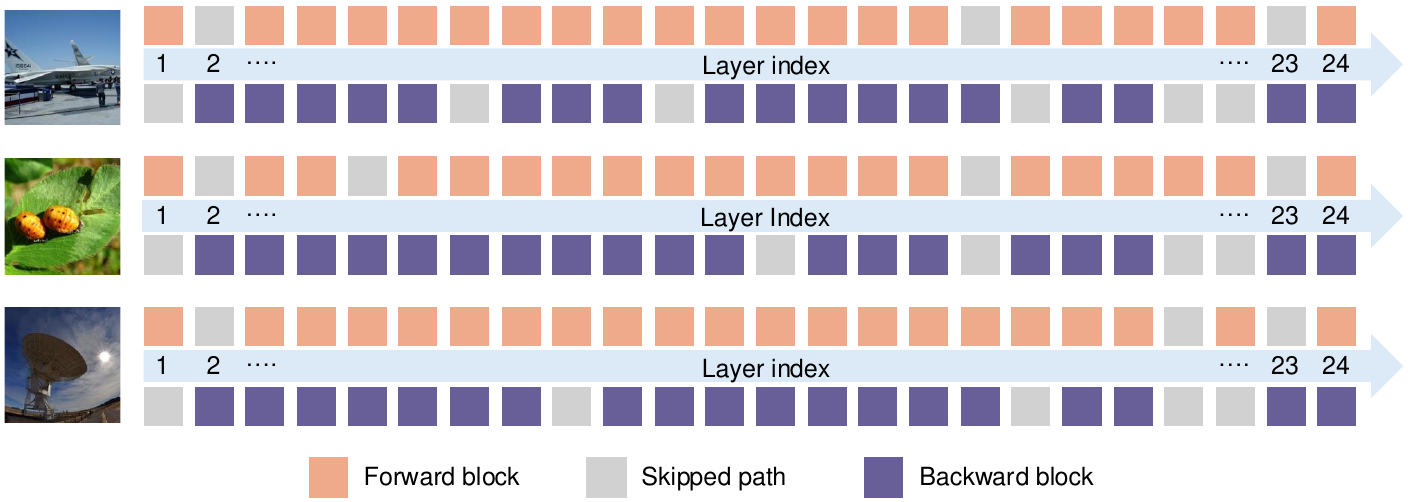}
    \caption{Visualization of the block selection policy. We present the policy at each layer with white squares denoting skipped blocks and colored ones denoting selected blocks. Different samples select different SSM block combinations across layers.}
    \label{blockVisual}
\end{figure*}

\section{Analysis}

\paragraph{Learnable token pruning and block selection.}
DyVM introduces learnable token and block score predictors for token pruning and block selection. We validate their effectiveness through ablation studies. For token pruning, we compare learnable predictors with two other pruning strategies: random selection and fixed positions (static). For block selection, since samples are free to go through any block(s), we only compare ours against random selection. As shown in Tables~\ref{tab:DifferentTokenPruningStrategy} and~\ref{tab:DifferentblockPruningStrategy}, learnable predictors achieve the highest accuracy by precisely identifying redundant tokens and blocks.

\paragraph{Stages of token pruning.}
In DyVM, the token pruning method employs a multi-stage approach.
An alternative approach could achieve the same final pruning rate using fewer stages but higher per-stage pruning rates.
Therefore, we conduct an ablation study to examine how the number of pruning stages affects model quality when the final pruning rate is fixed.
The results are reported in Table~\ref{tab:NumberOfPruningStage}, showing that higher accuracy can be achieved with more pruning stages.

\paragraph{Inputs of token mask predictors.}
DyVM's token mask predictor uses the token itself as input to decide retention or pruning. To evaluate the impact of different inputs, we analyze Mamba-generated variables ($\Delta$, $\bar{B}$, $C$ as in Section~\ref{preliminary}), which are input-dependent. An ablation study replaces the predictor's input with each variable individually. As shown in Table~\ref{tab:DifferentPredictor}, the predictor performs best with direct token input, suggesting that additional token transformations are unnecessary.

\paragraph{Effects of different losses.}
To verify the effect of different training losses, we experiment on the Vim-S model with distillation losses (Equation~\ref{eq:DisOut} and \ref{eq:DisToken}) removed and report the results in Table~\ref{tab:DiffererntLoss}. It shows that both losses boost model performance slightly. Two supervision losses (Equation~\ref{eq:token} and \ref{eq:block}) are not experimented with since they are crucial for controlling the pruning ratio.

\paragraph{Throughput.}
To evaluate whether our method improves the throughput of the model, we conduct tests on various devices.
The results, reported in Table~\ref{tab:throughput}, demonstrate that our method can achieve acceleration across all devices.
Notably, the improvement is more pronounced for larger models (\eg, Vim-B), which is consistent with FLOPs analysis in Table~\ref{tab:ImageClassification}.

\begin{table}[t]
    \centering
    \tablestyle{6pt}{1.2}
    \begin{tabular}{ccccc}
        model & \multicolumn{2}{c}{Vim-T} & \multicolumn{2}{c}{Vim-S} \\
        \shline
        predictor input & top-1 acc. & $\Delta_{\text{acc}}$ & top-1 acc. & $\Delta_{\text{acc}}$ \\
        $\Delta$ & 71.4 & \worseinv{2.3} & 78.0 & \worseinv{1.3} \\
        $\bar{B}$      & 71.3 & \worseinv{2.4} & 78.5 & \worseinv{0.8} \\
        $C$      & 71.5 & \worseinv{2.2} & 78.5 & \worseinv{0.8} \\ \rowcolor{gray}
        tokens   & 73.7 & / & \textbf{79.3} & / \\
    \end{tabular}
    \caption{Comparisons of different predictor input. Direct token input brings the best performance. We use token pruning with 0.7 target ratio and no block selection.}
    \vspace{-1em}
    \label{tab:DifferentPredictor}
\end{table}

\begin{table}[t]
    \centering
    \tablestyle{6pt}{1.2}
    \begin{tabular}{ccc}
         loss & top-1 acc. & $\Delta_{\text{acc}}$ \\
         \shline \rowcolor{gray}
         w/both & \textbf{78.8} & / \\
         w/o $\mathcal{L}_{dis\_token}$ & 78.5 & \worseinv{0.3} \\
         w/o $\mathcal{L}_{dis\_out}$ & 78.4 & \worseinv{0.4} \\
    \end{tabular}
    \caption{Effect of each loss. Both distillation losses contribute to model performance slightly. We report Vim-S accuracy with the same setting used in Table~\ref{tab:ImageClassification}.}
    \label{tab:DiffererntLoss}
\end{table}

\begin{table}[htb]
    \tablestyle{5pt}{1.2}
    \centering
    \begin{tabular}{ccccc}
        device       & V100 & A6000 & A100 \\ \shline 
        Vim-S        & 371.7 & 412.7     &  632.7    \\
        Vim-S + HA   & \better{27\%} & /        &  /       \\ \rowcolor{gray}
        Vim-S + DyVM & 497.9(\better{34.0\%}) & 602.4(\better{46.0\%})      & 823.7(\better{30.2\%})     \\ 
        \hshline
        Vim-B        & 147.3 & 174.5      & 238.5     \\ \rowcolor{gray}
        Vim-B + DyVM & 217.4(\better{47.6\%}) & 272.3(\better{56.0\%})      & 347.8(\better{45.8\%})    
        \end{tabular}
    \caption{DyVM consistently improves throughput of Vim-S and Vim-B on different devices. We test with the same pruned model in Table~\ref{tab:ImageClassification}, with 4 cards in parallel and total batch size of 1024.}
    \label{tab:throughput}
\end{table}

\paragraph{Visualization.}
We visualize predicted token pruning and block selection policies to demonstrate DyVM's efficacy. For token pruning, we show hidden attention heatmaps~\cite{ali2024hiddenattentionmambamodels} and retained tokens at each stage (Figure~\ref{TokenVisual}). Red regions denote high attention scores, while blue regions indicate low attention scores. Redundant tokens in inactive areas are pruned across stages, while discriminative features that receive high attention scores are retained. For block selection, policies vary across images (Figure~\ref{blockVisual}), highlighting DyVM's ability to customize paths for each sample. These visualizations underscore DyVM's effectiveness in reducing spatial redundancy in the Vim model.

\section{Conclusion}

In this work, we propose a novel method, DyVM, to improve the efficiency of Mamba-based vision models.
DyVM's rearranging strategy successfully resolves training-inference inconsistency with no extra computation overhead.
DyVM effectively reduces Vim's FLOPs and maintains comparable performance.  
We also make an early effort to reduce the number of scanning blocks and inspire future studies to maintain a good balance when designing new vision Mamba architectures.
In addition, DyVM demonstrates great generalization capability and improves the efficiency of other Mamba-based models in different vision tasks.

\paragraph{Acknowledgment}
This work was supported by Damo Academy through the Damo Academy Research Intern Program. This work was also supported by the National Research Foundation, Singapore under its AI Singapore Programme (AISG Award No: AISG2-PhD-2021-08-008). Yang You’s group is being sponsored by the NUS startup grant (Presidential Young Professorship), Singapore MOE Tier-1 grant, ByteDance grant, ARCTIC grant, SMI grant (WBS number: A8001104-00-00), Alibaba grant, and Google grant for TPU usage.

{
    \small
    \bibliographystyle{ieeenat_fullname}
    \bibliography{main}
}

\clearpage
\setcounter{page}{1}
\maketitlesupplementary

\appendix
\noindent We organize our supplementary material as follows:

\paragraph{Additional Experimental Results and Findings}

\begin{itemize}
\item Appendix~\ref{sec::ratio_combination}: Results of different token-block ratio combinations, as reported in Figure~\ref{fig:ratio-acc}.
\item Appendix~\ref{sec::throughput}: Throughput of different token-block ratio combinations.
\item Appendix~\ref{sec:pureTokenBlock}: Results of pure token selection or pure block selection.
\end{itemize}

\paragraph{Additional Visualization}

\begin{itemize}
    \item Appendix~\ref{sec::addtion_visualize}: More visualizations of token pruning.
\end{itemize}

\paragraph{Experimental Settings}
\begin{itemize}
    \item Appendix~\ref{sec::classification}: Experimental settings of image classification.
    \item Appendix~\ref{sec::video}: Experimental settings of video classification.
    \item Appendix~\ref{sec::semantic}: Experimental settings of semantic segmentation.
\end{itemize}

\section{Additional Experimental Results}
\label{sec::addition_results}
\subsection{Results of Different Ratio Combinations}
\label{sec::ratio_combination}
In Table~\ref{tab:DifferentRatioCombsTiny} and Table~\ref{tab:DifferentRatioCombsSmall}, we present detailed performances and GFLOPs of various token-block ratio combinations, complementing Figure~\ref{fig:ratio-acc}.
From the results presented in the tables, it can be observed that the Vim-S model exhibits greater pruning potential compared to the Vim-T model, with more pronounced spatial redundancy. Consequently, under smaller token-block ratio combinations, the performance degradation of Vim-S is less severe than that of Vim-T. Moreover, token pruning contributes more significantly to the reduction of FLOPs.

\begin{table*}[t]
\centering

    \begin{subtable}{1.0\textwidth}
        \tablestyle{15pt}{1.2}
        \begin{tabular}{c|llll}
        \multirow{2}{*}{Token Ratio} & \multicolumn{4}{c}{Block Ratio} \\
                                     & \multicolumn{1}{c}{0.6}    & \multicolumn{1}{c}{0.7}    & \multicolumn{1}{c}{0.8}   & \multicolumn{1}{c}{1.0}   \\ \shline
        0.7                          & 70.8(\worseinv{5.3})   & 71.5(\worseinv{4.6})   & 72.7(\worseinv{3.4})  & 73.4(\worseinv{2.7})  \\
        0.8                          & 72.9(\worseinv{3.2})   & 73.5(\worseinv{2.6})   & 74.0(\worseinv{2.1})  & 74.2(\worseinv{1.9})  \\
        0.9                          & 74.2(\worseinv{1.9})   & 74.7(\worseinv{1.4})   & 75.2(\worseinv{0.9})  & 75.5(\worseinv{0.6})  \\
        1.0                          & 74.5(\worseinv{1.6})   & 74.9(\worseinv{1.2})   & 75.6(\worseinv{0.5})  & 76.1 
        \end{tabular}
        \caption{Token pruning and dynamic block selection both have significant impacts on the performance of Vim-T. This shows that the spatial redundancies in Vim-T of both tokens and SSM blocks are smaller.}
    \end{subtable}
    
    \vfill
    
    \begin{subtable}{1.0\textwidth}
        \centering
        \tablestyle{15pt}{1.2}
        \begin{tabular}{c|llll}
        \multirow{2}{*}{Token Ratio} & \multicolumn{4}{c}{Block Ratio} \\
                                      & \multicolumn{1}{c}{0.6}    & \multicolumn{1}{c}{0.7}    & \multicolumn{1}{c}{0.8}   & \multicolumn{1}{c}{1.0}   \\ \shline
        0.7                          & 0.89(\betterinv{38.6\%})   & 0.90(\betterinv{37.9\%})   & 0.93(\betterinv{35.8\%})   & 0.96(\betterinv{34.1\%})     \\
        0.8                          & 1.02(\betterinv{30.0\%})   & 1.05(\betterinv{27.6\%})   & 1.08(\betterinv{25.7\%})   & 1.10(\betterinv{23.8\%})     \\
        0.9                          & 1.22(\betterinv{15.6\%})   & 1.24(\betterinv{14.4})   & 1.25(\betterinv{13.8\%})   & 1.28(\betterinv{11.8\%})     \\
        1.0                          & 1.38(\betterinv{4.8\%})    & 1.41(\betterinv{2.8\%})    & 1.42(\betterinv{2.1\%})    & 1.45 
        \end{tabular}
        \caption{Both token pruning and dynamic block selection contribute to FLOP reduction. Compared with block selection, the efficiency contribution of token pruning is more significant.}
    \end{subtable}
\caption{\textbf{(a)}: Performances of DyVM+Vim-T under different token-block ratio combinations on ImageNet-1K. \textbf{(b)}: GFLOPs of DyVM+Vim-T under different token-block ratio combinations on ImageNet-1K.}
\label{tab:DifferentRatioCombsTiny}
\end{table*}
    
\begin{table*}[t]
    \begin{subtable}{1.0\textwidth}
        \tablestyle{15pt}{1.2}
        \begin{tabular}{c|cccc}
        \multirow{2}{*}{Token Ratio} & \multicolumn{4}{c}{Block Ratio} \\
                                     & \multicolumn{1}{c}{0.6}    & \multicolumn{1}{c}{0.7}    & \multicolumn{1}{c}{0.8}   & \multicolumn{1}{c}{1.0}   \\ \shline
        0.7                          & 77.9(\worseinv{2.6})   & 78.4(\worseinv{2.1})   & 78.8(\worseinv{1.7})  & 78.8(\worseinv{1.7})  \\
        0.8                          & 79.2(\worseinv{1.3})   & 79.4(\worseinv{1.1})   & 79.5(\worseinv{1.0})  & 79.7(\worseinv{0.8})  \\
        0.9                          & 79.3(\worseinv{1.2})   & 79.6(\worseinv{0.9})   & 79.8(\worseinv{0.7})  & 80.3(\worseinv{0.2})  \\
        1.0                          & 79.5(\worseinv{1.0})   & 80.3(\worseinv{0.2})   & \textbf{80.5}  & 80.5 
        \end{tabular}
        \caption{ The pruning of tokens and blocks on Vim-S doesn't result in a remarkable performance drop, and selecting blocks only can achieve lossless performance. This implies that in Vim-S, spatial redundancies of image tokens and SSM blocks are substantial.}
    \end{subtable}

    \vfill
    
    \begin{subtable}{1.0\textwidth}
        \tablestyle{15pt}{1.2}
        \begin{tabular}{c|llll}
        \multirow{2}{*}{Token Ratio} & \multicolumn{4}{c}{Block Ratio} \\
                                     & \multicolumn{1}{c}{0.6}    & \multicolumn{1}{c}{0.7}    & \multicolumn{1}{c}{0.8}   & \multicolumn{1}{c}{1.0}   \\ \shline
        0.7                          & 3.2(\betterinv{37.0\%})   & 3.23(\betterinv{36.5\%})   & 3.29(\betterinv{35.2\%})   & 3.35(\betterinv{34.1\%})     \\
        0.8                          & 3.72(\betterinv{26.8\%})   & 3.75(\betterinv{26.2\%})   & 3.80(\betterinv{25.1\%})   & 3.87(\betterinv{23.8\%})     \\
        0.9                          & 4.30(\betterinv{15.4\%})   & 4.35(\betterinv{14.4\%})   & 4.42(\betterinv{13.0\%})   & 4.49(\betterinv{11.6\%})     \\
        1.0                          & 4.97(\betterinv{2.2\%})    & 5.05(\betterinv{0.6\%})  & 5.11(\worse{0.6\%})       & 5.08
        \end{tabular}
        \caption{Token pruning contributes to FLOP reduction more than dynamic block selection. Without token pruning, when the block ratio is too high, the reduced FLOPs cannot justify the added complexity by predictors.}
    \end{subtable}
\caption{\textbf{(a)}: Performances of DyVM + Vim-S under different token-block ratio combinations on ImageNet-1K. Bolded entries indicate lossless results. \textbf{(b)}: GFLOPs of DyVM + Vim-S under different token-block ratio combinations on ImageNet-1K.}
\label{tab:DifferentRatioCombsSmall}
\end{table*}

\begin{table*}[t]
    \tablestyle{15pt}{1.2}
    \begin{tabular}{cc|ccc}
    Model & Token Ratio & Accuracy & FLOPs \\
    \shline
    \multirow{4}{*}{Vim-T} & 1.0 & 76.1 & 1.45 \\
    & 0.9 & 75.5(\worseinv{0.6}) & 1.28(\betterinv{11.8\%}) \\
    & 0.8 & 74.2(\worseinv{1.9}) & 1.10(\betterinv{23.8\%}) \\
    & 0.7 & 73.4(\worseinv{2.7}) & 0.96(\betterinv{34.1\%}) \\
    \hshline
    \multirow{4}{*}{Vim-S} & 1.0 & 80.5 & 5.08 \\
    & 0.9 & 80.3(\worseinv{0.2}) & 4.49(\betterinv{11.6\%}) \\
    & 0.8 & 79.7(\worseinv{0.8}) & 3.87(\betterinv{23.8\%}) \\
    & 0.7 & 78.8(\worseinv{1.7}) & 3.35(\betterinv{34.1\%}) \\
    \end{tabular}
    \caption{The accuracy and throughput of pure token pruning with different token ratio, extracted from Table~\ref{tab:DifferentRatioCombsTiny} and \ref{tab:DifferentRatioCombsSmall}. Token pruning provides rapid FLOPs reduction, but with significant accuracy degradation.}
    \label{tab:pureToken}
\end{table*}

\begin{table*}[t]
    \tablestyle{15pt}{1.2}
    \begin{tabular}{cc|ccc}
    Model & Block Ratio & Accuracy & FLOPs \\
    \shline
    \multirow{4}{*}{Vim-T} & 1.0 & 76.1 & 1.45 \\
    & 0.8 & 75.6(\worseinv{0.5}) & 1.42(\betterinv{2.1\%}) \\
    & 0.7 & 74.9(\worseinv{1.2}) & 1.41(\betterinv{2.8\%})\\
    & 0.6 & 74.5(\worseinv{1.6}) & 1.38(\betterinv{4.8\%})\\
    \hshline
    \multirow{4}{*}{Vim-S} & 1.0 & 80.5 & 5.08 \\
    & 0.8 & 80.5 & 5.11(\worse{0.6\%}) \\
    & 0.7 & 80.3(\worseinv{0.2}) & 5.05(\betterinv{0.6\%}) \\
    & 0.6 & 79.5(\worseinv{1.0}) & 4.97(\betterinv{2.2\%}) \\
    \end{tabular}
    \caption{The accuracy and throughput of pure block pruning with different block ratio, extracted from Table~\ref{tab:DifferentRatioCombsTiny} and \ref{tab:DifferentRatioCombsSmall}. Block pruning provides slower accuracy degradation, but it also reduces less FLOPs.}
    \label{tab:pureBlock}
\end{table*}

\subsection{Throughput of Different Ratio Combinations}
\label{sec::throughput}
In the previous sections, we identified that the primary bottleneck in the throughput of the Vision Mamba model arises from the SSM blocks. To address this, we proposed the dynamic block selection mechanism to reduce redundancy in the block dimension. Table~\ref{tab:throughput_all} presents the throughput of DyVM applied to Vim-S under various token-to-block ratio configurations. As observed, the model's throughput improves significantly with a reduction in the block ratio, while a decrease in the token ratio has a less pronounced effect on throughput enhancement. This observation further supports our hypothesis and demonstrates the effectiveness of the proposed dynamic block selection module.

\begin{table*}[t]
    \tablestyle{19pt}{1.1}
    \begin{tabular}{c|ccc}
    \multirow{2}{*}{Token Ratio} & \multicolumn{3}{c}{Block Ratio} \\
                                 & \multicolumn{1}{c}{0.6}    & \multicolumn{1}{c}{0.7}    & \multicolumn{1}{c}{0.8}   \\ \shline
    0.7                          &
    950.16(\better{24.9\%})   & 
    896.66	(\better{17.9\%}) & 886.55(\better{16.6\%})    \\
    0.8                          & 
    966.86(\better{27.1\%})   &	
    904.60(\better{19.0\%})  &
    853.23(\better{12.2\%})    \\
    0.9                          &
    931.40(\better{22.5\%})   & 
    922.72(\better{21.3\%})   & 	893.79(\better{17.5\%})    
    \end{tabular}
\caption{The throughput of DyVM + Vim-S under different token-block ratio combinations on ImageNet-1K. All results are obtained on NVIDIA A100.}
\label{tab:throughput_all}
\end{table*}

\subsection{Results of Pure Token or Block Pruning}
\label{sec:pureTokenBlock}

We extract the pure token pruning and pure block pruning data from Table~\ref{tab:DifferentRatioCombsTiny} and \ref{tab:DifferentRatioCombsSmall} to better understand the effect of each pruning module. The results are reported in Table~\ref{tab:pureToken} and \ref{tab:pureBlock}, respectively. We can observe that while token selection reduces FLOPs more rapidly, it also degrades accuracy significantly. Block selection, on the other hand, degrades accuracy slower but also provides less FLOPs reduction under the same ratio setting.

\section{Additional Visualization}
\label{sec::addtion_visualize}
\begin{figure*}
    \centering
    \includegraphics[width=1.0\linewidth]{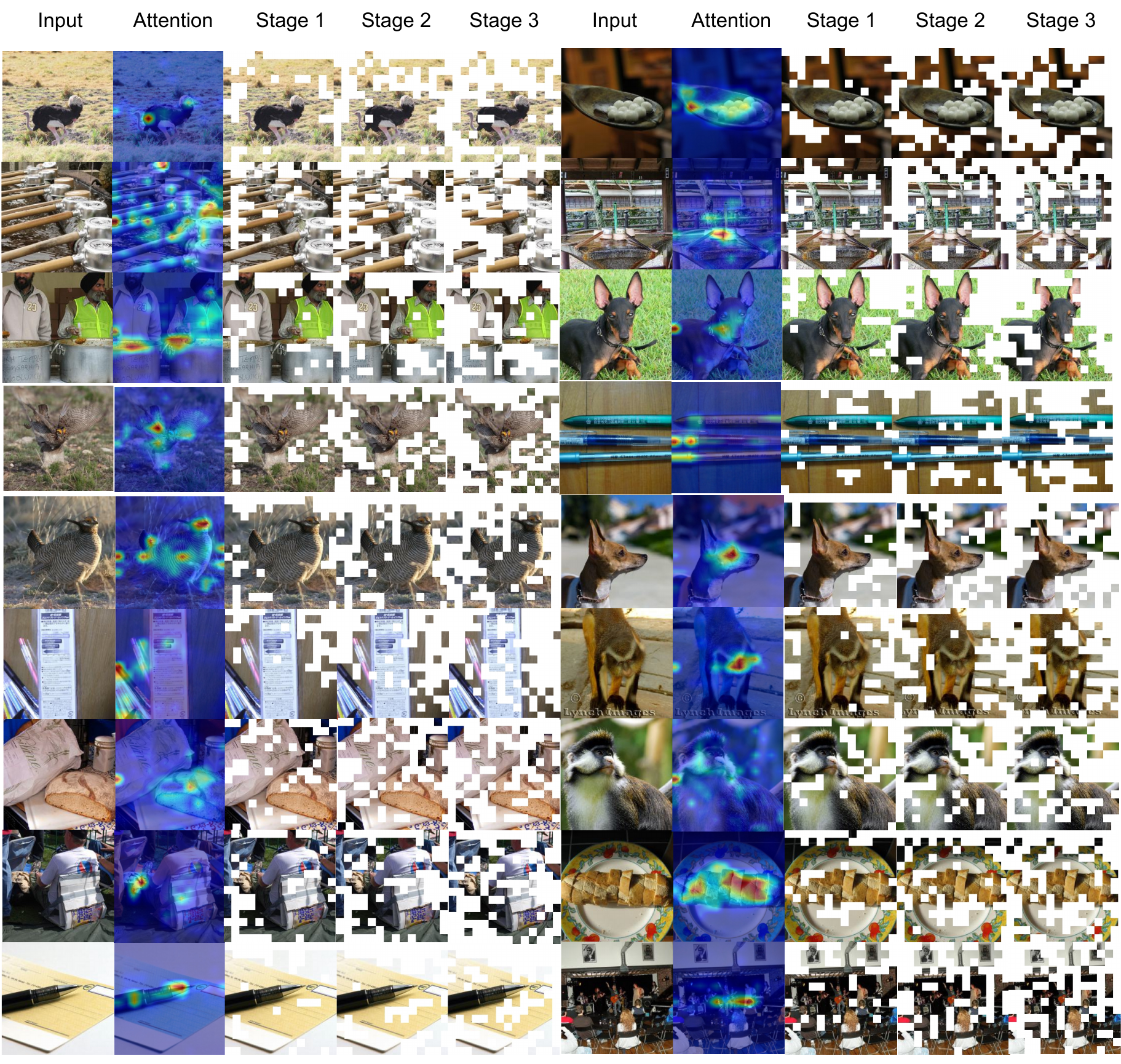}
    \caption{Visualization of token pruning results. In each group of images, we show the original image, along with its hidden attention and retained tokens of each pruning stage. Pruned tokens are mostly from low-attention areas, implying their redundancy. (In the attention figure, red zone represents high attention score and blue zone represents low attention score)}
    \label{fig::supple_visualization}
\end{figure*}

In this section, we provide additional visualization of our results shown in Figure~\ref{fig::supple_visualization}. We can observe that Vim can extract vision features effectively and use these features to carry out image classification. Moreover, our Dyvm manages to preserve most of the tokens containing rich information while pruning other redundant tokens. This serves as a solid indicator that our method strikes a balance between efficiency and accuracy.

\section{Experimental Settings}

\subsection{Image Classification}
\label{sec::classification}
In Table~\ref{tab:hyper-param}, we show hyper-parameters of the experiments done in Table~\ref{tab:ImageClassification}.
We split them into two modules: \textit{DyVM} includes main hyper-parameters of token pruning and dynamic block selection, and \textit{Training} includes main hyper-parameters of the objective function and optimization.
For ablations on different token-block ratio combinations, we follow the same training recipe except that $\lambda_{token}$ and $\lambda_{block}$ are both set to 10 for Vim-T to ensure the block selection fits the target ratio better. Our results are not sensitive to hyper-parameters.
\begin{table*}[t]
\tablestyle{12pt}{1.2}
\centering
\begin{tabular}{c|ccc|cccccc}
Modules          & \multicolumn{3}{c|}{DyVM}     & \multicolumn{6}{c}{Training}                                                                       \\
Hyper-parameters & stage       & $\rho^S$ & $\rho^P$ & $\lambda_{cls}$ & $\lambda_{token}$ & $\lambda_{block}$ & $\lambda_{dis\_out}$ & $\lambda_{dis\_token}$ & $lr$   \\ \shline
Vim-T            & [6, 12, 18] & 0.9    & 0.8    & 1.0           & 10.0            & 10.0            & 0.5               & 0.5                 & 3e-5 \\
Vim-S            & [6, 12, 18] & 0.7    & 0.8    & 1.0           & 10.0            & 10.0            & 0.5               & 0.5                 & 5e-5 \\
Vim-B            & [6, 12, 18] & 0.7    & 0.7    & 1.0           & 10.0            & 10.0             & 0.5               & 0.5                 & 5e-5 
\end{tabular}
\caption{Hyper-parameters of experiments conducted in Table~\ref{tab:ImageClassification}. For VideoMamba and MambaReg, we use the same settings as Vim for each corresponding model size.}
\label{tab:hyper-param}
\end{table*}

\subsection{Video Classification}
\label{sec::video}
For video classification, we adapt from the official 8-frame version training script of VideoMamba~\cite{VideoMamba}. We pruned at layer 6, 12 and 18, and set the target token ratio to 0.7 and block ratio to 0.8.

\subsection{Semantic Segmentation}
\label{sec::semantic}

Since the dense prediction task demands a complete feature map, DyVM is not directly applicable to semantic segmentation tasks as redundant tokens are pruned. To address this, instead of dropping the pruned tokens, we remove them from sequence and stop their updates. After the forwarding process, we restore the pruned tokens and retained tokens to their original positions in the feature map.

We use the same training setting as Vim~\cite{Vim}: we use UperNet~\cite{xiao2018unifiedperceptualparsingscene} as the framework and our DyVM model as the backbone. In training, we employ AdamW with a weight decay of 0.01, and a total batch size of 16 to optimize models. The employed training schedule uses an initial learning rate of 6e-5, linear learning rate decay, a linear warmup of 1500 iterations, and a total training of 160K iterations. We pruned at layer 6, 12 and 18. We set target token ratio to 0.9 for Vim-T and 0.8 for Vim-S. The block ratio is set to 0.8 for both models.

\end{document}